\documentclass[runningheads]{llncs}
\usepackage{graphicx}
\graphicspath{{./}}
\usepackage[listofformat=parens, justification=centering, subrefformat=subparens]{subfig}

\usepackage{amsmath,amssymb} 
\usepackage{color}
\usepackage[width=122mm,left=12mm,paperwidth=146mm,height=193mm,top=12mm,paperheight=217mm]{geometry}
\usepackage{comment}
\usepackage{caption}
\usepackage{enumitem}

\usepackage{pdfpages}
\usepackage{fancyhdr}

\DeclareMathOperator*{\argmin}{\arg\!\min}
\newcommand\expected{\mathbb{E}}

\begin{document}
\pagestyle{headings}
\mainmatter
\def\ECCV16SubNumber{875}  

\title{\textit{MOON}: A \textit{Mixed Objective Optimization Network} for the Recognition of Facial Attributes} 

\titlerunning{\textit{MOON}: A \textit{Mixed Objective Optimization Network}  }

\authorrunning{Ethan M. Rudd, Manuel G\"unther, and Terrance E. Boult}

\author{Ethan M. Rudd, Manuel G\"unther, and Terrance E. Boult}


\institute{Vision and Security Technology (VAST) Lab,\\
        University of Colorado at Colorado Springs\\
        \email{ \{erudd,mgunther,tboult\}@vast.uccs.edu}
}


\maketitle

{
  \chead{\footnotesize This is a post-print of the original paper published in ECCV 2016 (\href{http://link.springer.com/chapter/10.1007\%2F978-3-319-46454-1_2}{SpringerLink}).}
  \lhead{}
  \thispagestyle{fancy}
}

\begin{abstract}
Attribute recognition, particularly facial, extracts many labels for each image.
While some multi-task vision problems can be decomposed into separate tasks and stages, e.g., training independent models for each task, for a growing set of problems joint optimization across all tasks has been shown to improve performance.
We show that for deep convolutional neural network (DCNN) facial attribute extraction, multi-task optimization is better.
Unfortunately, it can be difficult to apply joint optimization to DCNNs when training data is imbalanced, and re-balancing multi-label data directly is structurally infeasible, since adding/removing data to balance one label will change the sampling of the other labels.
This paper addresses the multi-label imbalance problem by introducing a novel mixed objective optimization network (MOON) with a loss function that mixes multiple task objectives with domain adaptive re-weighting of propagated loss.
Experiments demonstrate that not only does MOON advance the state of the art in facial attribute recognition, but it also outperforms independently trained DCNNs using the same data.
When using  facial attributes for the LFW face recognition task,   we show that our balanced (domain adapted) network outperforms the unbalanced trained network.

\end{abstract}



\renewcommand\cap[3]{\caption[#2]{\label{#1}\textsc{#2}. \small\textit{#3}}}

\section{Introduction}
Given an input image or video, there are often multiple vision tasks to be accomplished, i.e., multiple objectives to be optimized.
Under certain constraints, e.g., when tasks feed into each other, or when there is need to share computed features or representations, then multiple task objectives can benefit from being mixed and jointly optimized.
This kind of multi-objective learning has affected many areas of computer vision including scene/object classification and annotation \cite{zhou2007multi,quattoni2008transfer,huang2013multi,wu2015weakly}, tracking \cite{zhang2012robust}, facial landmark estimation \cite{zhang2014facial,zhang2014improving}, face verification \cite{wang2009boosted}, and face detection with head pose estimation \cite{yan2013no,ouyang2014multi,yim2015rotating}.

\begin{figure*}[t!]
  \begin{center} \includegraphics[width=.99\textwidth]{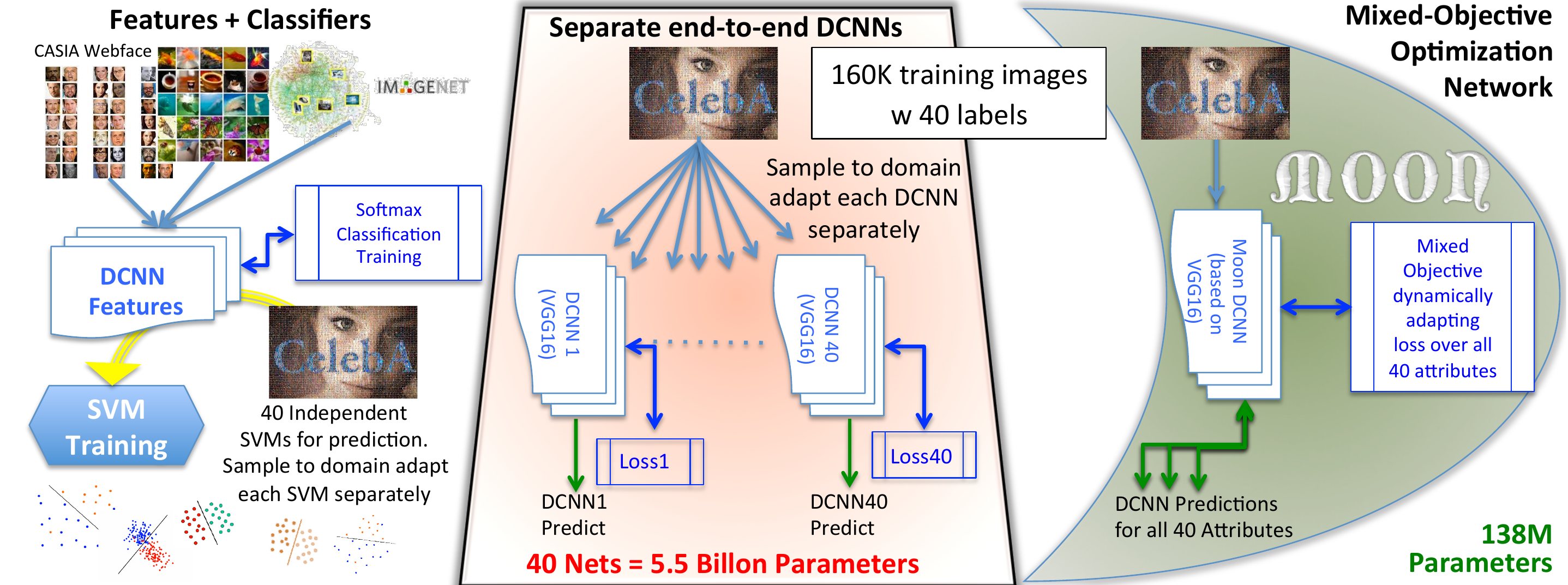} \end{center}  
\cap{fig:teaser}{Three approaches to attribute learning (and other multi-task problems)}{\small  In the left is a conceptual model of previous state-of-the-art approaches, with features trained for classification problems and then adapted as inputs to independent SVMs for prediction. The middle approach attacks the problem with separately trained deep convolution neural networks (DCNNs). While we demonstrate that this advances the state of the art in attribute accuracy, it is cost prohibitive. This paper shows that for attributes, joint multi-task learning does better.  However, for multi-label learning there is no way simply to re-weight or sample inputs to deal with imbalance or domain adaption because each input defines values for all attributes.  On the right is our answer, the mixed objective optimization network (MOON) architecture with a domain adaptive multi-task DCNN loss. To adapt, for each input the MOON objective re-weights each part of the loss associated with each attribute. MOON learns to balance its multi-task output predictions with reduced training and storage costs, while producing better accuracy than independently trained DCNNs.} 
\end{figure*}

This paper addresses facial attribute recognition, which we hypothesize is well suited to a multi-objective approach because facial attributes have a shared, albeit latent correlation that imposes soft constraints on the space of attributes, e.g., $p(\texttt{Male} | \texttt{Mustache}) \approxeq 1$.
Despite the fact that facial attribute recognition inherently seeks multiple labels for the same image, multi-objective learning has not been widely applied to facial attributes.
One potential reason is that balancing the training for the labels is difficult.
Prior approaches to facial attribute recognition independently optimize a choice of features and recognition model (Features+Classifiers in Fig.~\ref{fig:teaser}). For example, the original approach taken by Kumar et al.
~\cite{kumar2008facetracer} used AdaBoost to select a separate feature space for each attribute and independent SVMs to perform classification.
Likewise, the current state of the art \cite{liu2015deep} trains DCNN features with facial identity recognition and localization datasets and then trains independent SVMs in this feature space for attribute classification.
In both cases the separation makes it easy to re-balance training per attribute.

In this work, we show that a joint optimization with respect to all attributes offers performance superior to the state-of-the-art Features+Classifiers approach.
We also show that joint optimization over all attributes outperforms training a single independent network of similar topology per attribute, in which the feature space is optimized along with the classifier on a per-attribute basis, both in terms of accuracy and storage/processing efficiency.
This result suggests that the multi-task approach is far more effective at distilling latent correlations than relying on independent classifiers to learn them implicitly.
Thus, not only is a multi-objective approach far more intuitive, it is also far more effective.

It is unlikely that the source distribution of binary facial attributes for the training set will match the target distribution of the test set.
We would like facial attribute classifiers trained on a dataset of one demographic to still work well for discrimination on a different demographic; thus some sort of domain adaptation is required.
Many approaches to domain adaption exist \cite{patel2015visual}, with input sampling or re-weighting being common.
Unfortunately, multi-objective training introduces challenges because balanced training is difficult or impossible via input sampling or weighting.
Given a target distribution, domain adaptation is easy for separately trained attribute classifiers, e.g., by re-weighting errors in the cost function in each classifier.
However, it is less immediately obvious how to do this in training for a multi-objective classifier.
To this end, we introduce the MOON (Mixed-Objective Optimization Network) architecture.
MOON is a novel multi-objective neural network architecture, which {\em mixes} the tasks of multi-label classification and domain adaptation under one unified objective function.

In summary, the contributions of this paper include:
\begin{itemize}[noitemsep,topsep=0pt]
\item{ A  mixed objective optimization network (MOON) architecture, which advances face attribute recognition by learning  multiple attribute labels simultaneously via a single DCNN that supports domain adaption for multi-task DCNNs.}
\item{A fair evaluation technique which incorporates source and target distributions into the classification measure, leading to the balanced CelebA (CelebAB) evaluation protocol,}
\item{Experiments demonstrating that the MOON architecture significantly advances state-of-the-art attribute recognition on the CelebA dataset, improving both accuracy and efficiency. These experiments also demonstrate that optimizing over all attributes simultaneously offers a noticeable reduction in classification error compared to optimizing single attributes over the same dataset and network topology.}
\item{Experiments showing that domain adaptation on attribute classifiers trained on CelebA  enhances the recognition capacity of MOON attributes on LFW, advancing attribute-based face recognition. }
\item{Evaluation of stability of the MOON architecture to fiducial perturbations and data set imbalance.}
\end{itemize}

\section{Related Work}
Multi-task learning has been applied to several areas that rely on learning fine-grained discriminations or localizations under the constraint of a global correlating structure.
In these problems, multiple target labels or objective functions must simultaneously be optimized.
In object recognition problems, multiple objects may be present in a training image whose co-occurrences should be explicitly learnt~\cite{wei2014cnn}.
In text classification problems, joint inference across all characters in a word yields performance gains over independent classification~\cite{jaderberg2014deep}.
In multi-label image tagging/retrieval~\cite{wu2015weakly,huang2015unconstrained}, representations of the contents of an image across modalities (e.g., textual descriptions, voice descriptions) are jointly inferred from the images.
The resulting classifiers can then be used to generate descriptions of novel images (tagging) or to query images based on their descriptions (retrieval).
Closer to this work, facial model fitting and landmark estimation~\cite{cootes2001active,blanz2003face} is another multi-task problem, which requires a fine-grained fit due to tremendous diversity in facial features, poses, lighting conditions, expressions, and many other exogenous factors.
Solutions also benefit from global information about the space of face shapes and textures under different conditions. Optimization with respect to local gradients and textures is necessary for a precise fit, while considering the relative locations of all points is important to avoid violating facial topologies.

This paper applies multi-task learning to facial attributes.
Applications of facial attributes include searches based on semantically meaningful descriptions (e.g., ``Caucasian female with blond hair'')~\cite{kumar2008facetracer,kumar2011describable,scheirer2012multi}, verification systems that explain in a human-comprehensible form \textit{why} verification succeeded or failed~\cite{kumar2009attribute}, relative relations among attributes \cite{parikh2011interactively}, social relation/sentiment analysis~\cite{zhang2015learning}, and demographic profiling.
Facial attributes also provide information that is more or less independent of that distilled by conventional recognition algorithms, potentially allowing for the creation of more accurate and robust systems, narrowing down search spaces, and increasing efficiency at match time.

The classification of facial attributes was first pioneered by Kumar et al.
\cite{kumar2009attribute}.
Their classifiers depended heavily on face alignment, with respect to a frontal template, with each attribute using AdaBoost-learnt combinations of features from hand-picked facial regions (e.g., cheeks, mouth, etc.).
The feature spaces were simplistic by today's standards, consisting of various normalizations and aggregations of color spaces and image gradients.
Different features were learnt for each attribute, and a single RBF-SVM per attribute was independently trained for classification.
Although novel, the approach was cumbersome due to high dimensional varying length features for each attribute, leading to inefficiencies in feature extraction and classification \cite{wilber2014exemplar}.

In recent years, approaches have been developed to leverage more sophisticated feature spaces.
For example, gated CNNs~\cite{kang2015face} use cross-correlation across an aligned training set to determine which areas of the face are \textit{most relevant} to particular attributes.
The outputs of an ensemble of CNNs, one trained for each of the relevant regions, are then joined together into a global feature vector.
Final classification is performed via independent binary linear SVMs.
Zhang et al.~\cite{zhang2015learning} use CNNs to learn facial attributes, with the ultimate goal of using these features as part of an intermediate representation for a Siamese network to infer social relations between pairs of identities within an image.
Liu et al.~\cite{liu2015deep} use three CNNs -- a combination of two \textit{localization networks} (LNets), and an \textit{attribute recognition network} (ANet) to first localize faces and then classify facial attributes in the wild.
The localization network proposes locations of face images, while the attribute network is trained on face identities and attributes, and is used to extract features, which are fed to independent linear SVMs for final attribute classification.
Their approach was the state-of-the-art on the CelebA dataset at the time of the submission of this paper -- and serves as a basis of comparison.  In contrast to our approach, Liu et al. and many other recent works do not directly use attribute data in learning a feature space representation, but instead use truncated networks trained for other tasks.
While research suggests that coarse-grained attribute data (e.g., image-level) can be indirectly embedded into the hidden layers of large-scale identification networks \cite{escorcia2015relationship}, the efficiency of this approach has not been well studied for inferring fine-grained (e.g., facial) attribute representations, and findings from \cite{zhong2016leveraging} suggest that optimal implicit representations reside across different layers depending on the attribute.

Surprisingly, multi-task learning has not been widely applied to the problem of facial attribute recognition.
Only very recently has it been addressed, e.g., Ehrlich et al.~\cite{ehrlich2016facial} developed a Multi-Task Restricted Boltzmann Machine (MT-RBM).
In terms of joint inference for facial attributes, it is the first we could find in the literature, but the approach deviates radically from DCNN approaches in many other respects as well: the MT-RBM is generative and non-convolutional and it is unclear what contributed most to their improvement over \cite{liu2015deep}.

While there has been significant prior work in visual domain adaptation \cite{patel2015visual}, including more recent work for CNNs \cite{tzeng2015simultaneous}, the main problem that we address in this paper -- incorporating domain adaptation into the training procedure for multi-objective attribute classifiers -- has heretofore not been addressed, either in DCNN multi-task learning or in facial attribute research.
For facial attributes in particular, we contend that domain adaptation is essential when building classifiers fit to chosen target demographics.
Recently, Wang et al.~\cite{wang2016walk} demonstrated that even throughout New York City, a relatively compact geographic region, differences in demographic profile are so prominent as a function of geolocation that binned geolocation can be used to derive a powerful unsupervised facial attribute feature space representation.
In order to leverage attribute data we have for training demographic-specific classifiers, domain adaptation during training is vital to provide a balanced representation and mitigate problems from an over-correlated representation \cite{jayaraman2014decorrelating}.

\section{Approach}
\label{sec:approach}
\def\expected{\mathbb{E}}

For multi-task problems, the high level goal is to maximize accuracy over all tasks, where each task has its own objective. In our case, the task is attribute prediction, and we seek to simultaneously maximize prediction accuracy over all attributes.

Formally, let $\mathbb{I}$ be the space of allowable images, and let $M$ be the number of attributes. For a given sample $x\in \mathbb{I}$, let $y_i \in \{-1,+1\}$ be the binary ground truth label for $x$'s $i$th attribute, where $i \in \{1,\ldots,M\}$ is the attribute index. Let ${\cal H}$ be the space of allowable decision functions and $f_i(x;\theta_i)\in {\cal H}$ be the decision function, with parameters $\theta_i$, learnt for the $i$th attribute classifier. Given a set of loss functions $L_i(f_i(x;\theta_i),y_i)$, each of which defines the cost of an error on input $x$ with respect to attribute $i$,  let $\expected(f_i(x;\theta_i),y_i)$ be the expected value of that loss over the range of inputs $\mathbb{I}$.
Then the idealized problem is to minimize the loss for each attribute, i.e.:
\begin{equation}
\forall i\colon f^*_i = \argmin_{f_i\in{\cal H}} \expected(f_i(x;\theta_i),y_i).
\end{equation}
For input $x$ and attribute $i$, the classification  result $c_i(x)$ and its corresponding error $ e_i(x,y_i)$ are obtained by thresholding the associated prediction:
\begin{equation}
    c_i(x) = \begin{cases} +1 & \text{if } f_i(x) > 0 \\ -1 & \text{otherwise,}\end{cases}
     \ \text{and} \quad
    e_i(x,y_i) = \begin{cases} 0 & \text{if } y_ic_i(x) > 0 \\ +1 & \text{otherwise.}\end{cases}
    \label{eq:classification}
\end{equation}

Intuitively,  this appears to lead to $M$ independent optimization problems, for which one should be able to optimize each $f_i$ separately.
Accordingly, the most common approach to attribute classification in prior work is to use independent binary classifiers in some characteristic feature space to classify each attribute~\cite{kumar2009attribute,liu2015deep}. 
Both approaches in ~\cite{kumar2009attribute} and \cite{liu2015deep} learn $M$ independent binary classifiers trained with a hinge-loss objective.  The hinge-loss objective function is:
\begin{equation}
\label{eq:hinge_loss}
\argmin_{\theta_i} L_i(x,\theta_i,y_i) = max(0,1-y_if_i(x;\theta_i)).
\end{equation}
When the classifier is a dot product, i.e., $f_i(x) = \theta_i^{T}(1,x^T)^T$, solving this objective function results in a binary \textit{support vector machine} (SVM) -- the hyperplane that separates the two binary classes of data ($+1$ and $-1$) with maximum soft-margin. Given $M$ attributes, this approach leads to $M$ binary classifiers, each of which outputs a decision score. A positive decision score corresponds to the predicted presence of an attribute, while a negative decision score corresponds to its absence.


In order to learn latent correlations, it is also important to use attribute data directly to derive the feature space. Although Liu et al. \cite{liu2015deep} claim that latent features of attributes are learnt by their feature space representation while optimizing over a dataset for an identification task, the extent to which this is true for attributes that have little to do with identity (e.g., \texttt{Smiling}) is questionable. Rather, intuition suggests the opposite -- that networks trained for identification of individuals would learn to ignore such attributes. To uncover such correlations, the network used to learn the feature space should be directly trained on attribute data and the distribution of attributes in training should match the operational or testing distribution.

This leads to the problem of how to appropriately balance the dataset used to learn attribute features. A perfectly balanced dataset can be obtained by collecting separate images for each attribute, but this leads to an enormous dataset, with different identities for different attributes, effectively yielding a relatively small number of training images per attribute in proportion to the size of the dataset ~\cite{kumar2009attribute}. This approach also does not account for label correlations. Using a multi-label dataset, e.g., CelebA~\cite{liu2015deep} allows us to leverage multiple labels in a mixed objective, but the distribution is highly imbalanced for many attributes (cf. Sec.~\ref{sec:experiments}). Unfortunately, the attribute distribution of a given target population does not always follow the dataset bias.

In a separate per-class training,  balancing the number of positive and negative examples that are input to the classifier is easy, e.g., by weighting or sampling.
However, input balancing is nearly impossible for multi-task training.
Furthermore, for many tasks, the training frequencies and the operational/test frequencies will not match.
Our solution to both problems is to define a mixed objective function including domain adapted weights that incorporate the difference between the source and target distributions.
First, we compute the source distribution $S_i$ from the training set for each attribute $i$ by counting the relative number of occurrences of positive $S_i^+$ and negative samples $S_i^-$.
Given a binary target distribution, $T_i^+$ and $T_i^-$, for each attribute $i$ we assign a probability for each class:
\begin{equation}
  p(i|+1) =
    \begin{cases}
      1 & \text{if } T_i^+ > S_i^+ \\[1ex]
      \frac{S_i^- T_i^+}{S_i^+ T_i^-} & \text{otherwise}
   \end{cases}
   \ \text{and}\quad
  p(i|-1) =
    \begin{cases}
      1 & \text{if } T_i^- > S_i^- \\[1ex]
      \frac{S_i^+ T_i^-}{S_i^- T_i^+} & \text{otherwise.}
   \end{cases}
\end{equation}

We would like to incorporate this domain adaptation directly into a loss function, but we need a loss function that additionally mixes all attribute predictions and simultaneously infers latent correlations between attribute labels and image data. One approach would be to combine all of the objective functions for each attribute into one joint objective function, e.g.:
\begin{equation}
\label{eq:combined_objective}
\argmin_{\theta} \sum_{i=1}^{M} L_i(x, \theta, y_i),
\end{equation}
where $\theta$ are the parameters of the joint classifier, which for legibility reasons we omit from the following equations. We can then solve that optimization problem via backpropagation using raw attribute images and labels as a training set.
While we could use many potential loss functions, in our formulation we optimize a weighted mixed task squared error. Let $M$ be the number of attributes, ${\bf X}$ be a data tensor containing $N$ input images, and ${\bf Y}$ be a corresponding $N \times M$ matrix of labels. Then our domain-adapted multitask loss function is given by: 
\begin{equation}
  \label{eq:dataset_error}
  L({\bf X},{\bf Y}) = \sum_{j=1}^N \sum_{i=1}^{M} p(i | Y_{ji})\ {|| f_i(X_j) - Y_{ji} ||}^{2}.
\end{equation}
Replacing the standard loss layer of a DCNN with a layer implementing Eq.~\eqref{eq:dataset_error} results in the \textit{mixed objective optimization network} (MOON) architecture, which incorporates attribute correlations and can adapt the bias of the training dataset to a target distribution.
In our custom implementation we obtain the weights $p(i | Y_{ji})$ via sampling.
For each attribute $i$ with target value $Y_{ji} \in \{-1,+1\}$ we only backpropagate the error with the probability $p(i|Y_{ji})$, otherwise we set the gradient for attribute $i$ to 0.
The more source and target distributions differ, the more elements in the gradient are reset.

\section{Experiments}
\label{sec:experiments}

\subsection{Dataset}

\begin{figure*}[t!]
  \centering\includegraphics[width=.95\linewidth]{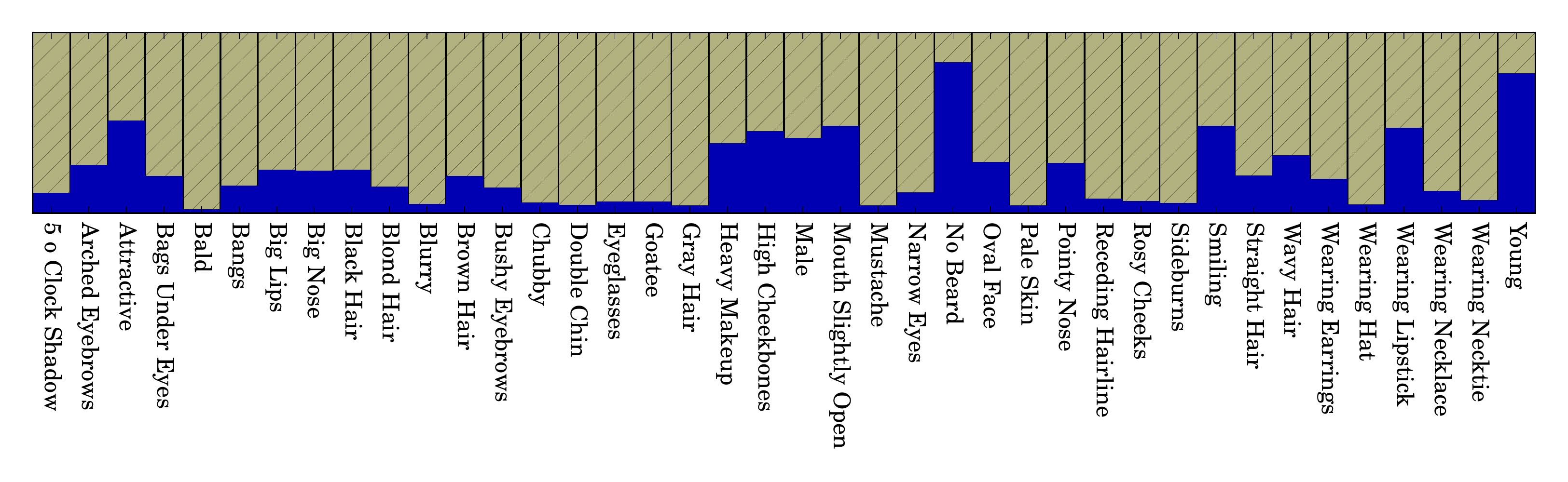}
  \cap{fig:DatasetBias}{CelebA Dataset Bias}{This figure shows the distribution of the attribute labels throughout the CelebA dataset: presence (blue) or absence (tan). }
\end{figure*}

For comparison with other attribute benchmarks, we conducted our experiments on the CelebA dataset \cite{liu2015deep}.
The dataset consists of batches of 20 images from approximately 10K celebrities, resulting in a total of more than 200K images.
Following the standard CelebA evaluation protocol, 8K identities (160K images) are used for training, 1K for validation and 1K for testing.
Each image is annotated with 5 key points (both eyes, the mouth corners and the nose tip), as well as binary labels of 40 attributes.
These attributes are shown in Fig.~\ref{fig:DatasetBias}, which also shows the relative number of images in which the attribute is hand-labeled as present (blue) or absent (tan), respectively. As one can observe, for many of the attributes, there is a strong bias for either of the two classes. This is especially the case for certain attributes, e.g., relatively few images are labeled as \texttt{Bald} or \texttt{Wearing Hat}, while the majority of the facial images are labeled as \texttt{Young}.

The CelebA dataset provides a set of pre-cropped face images, which were aligned using the hand-labeled key points.
For our experiments we use these images, but later (cf. Sec.~\ref{sec:misalignment}) we show that the trained classifier can also work with faces which are not perfectly aligned, and we introduce ideas to make our MOON network more robust to mis-alignment.


\subsection{Evaluating MOON on CelebA}
\label{sec:celebA}
In order to compare with existing approaches, which do not account for dataset bias, we evaluate MOON on the CelebA dataset, setting the target distribution to the source distribution, i.e., $\forall i\  T_i \equiv S_i$.

Using the CelebA training set, we trained a DCNN to predict attributes under a MOON architecture.
As the basic network configuration, we adopted the 16 layer VGG network from \cite{parkhi2015deep}, where we replaced the final loss layer with the loss in Eq.~\eqref{eq:dataset_error}.
We also changed the dimension of the RGB image input layer from $224\times224$ pixels to $178\times 218$ pixels, the resolution of the aligned CelebA images.
In opposition to \cite{parkhi2015deep}, we do not incorporate any dataset augmentation or mirroring, but train the network purely on the aligned images.
Due to memory limitations, the batch size was set to 64 images per training iteration and, hence, the training requires approximately 2500 iterations to run a full epoch on the training set.
We selected a learning rate of 0.00001, finding empirically that higher learning rates caused the network to learn only the bias of the training set.
During training we update the convolution kernel weights using the backpropagation algorithm with an \textit{RMSProp} update rule and an inverse learning rate decay policy.

We ran two types of network training, one training a separate network for each attribute, and one optimizing the combined MOON network.
Separately training classifiers is the most common approach taken in the literature.
By training one network per individual attribute, each network can concentrate only on the parts of the image it deems relevant to that attribute. During the separate training, we presented each network with all images from the training set, and a single input to the loss layer encoded with labels that denoted the presence ($+1$) or the absence ($-1$) of the attribute. Loss was computed according to Eq.~\eqref{eq:dataset_error}.
As each network required several hours to train on an NVIDIA Titan-X GPU, we chose to train each network for $\approx2$ epochs (5000 iterations).
To check if 2 epochs are sufficient to attain convergence to a maximum validation accuracy, we continued training for four attributes.
We selected these attributes -- \texttt{Attractive}, \texttt{Chubby}, \texttt{Narrow Eyes}, and \texttt{Young} -- to have varying statistics from the dataset: While \texttt{Attractive} is relatively balanced, images with \texttt{Chubby} and \texttt{Narrow Eyes} are mostly absent from the dataset, whereas \texttt{Young} is over-represented.
While errors on the training set further decreased, errors on the validation set \textit{increased} after approximately 4 - 6 epochs, with little improvement over the 2 epochs networks. This leads us to believe that improvements in validation accuracy beyond 2 epochs are negligible.

When training our MOON network, we use a single network with $M=40$ outputs to learn all attributes simultaneously. 
Since CelebA has identical source and target distributions, we define the loss layer in \eqref{eq:dataset_error} to weight all elements equally during backpropagation -- which is equivalent to Euclidean loss between the network output and the 40 binary attribute values.
We trained the network for 40 epochs since the validation error after 10 epochs was still decreasing.
Based on the minimum validation set error, we chose our final MOON network after 24 epochs.
While individual classifiers seem to take fewer training iterations than MOON to minimize their validation error,  the total training time of the MOON network is still lower than the sum of the separate network training times.
We suspect that the additional iterations required for the MOON network to converge are needed to learn a more sophisticated latent structure than those learnt by the separate networks.

\begin{figure*}[!t]
  \centering\includegraphics[width=\linewidth]{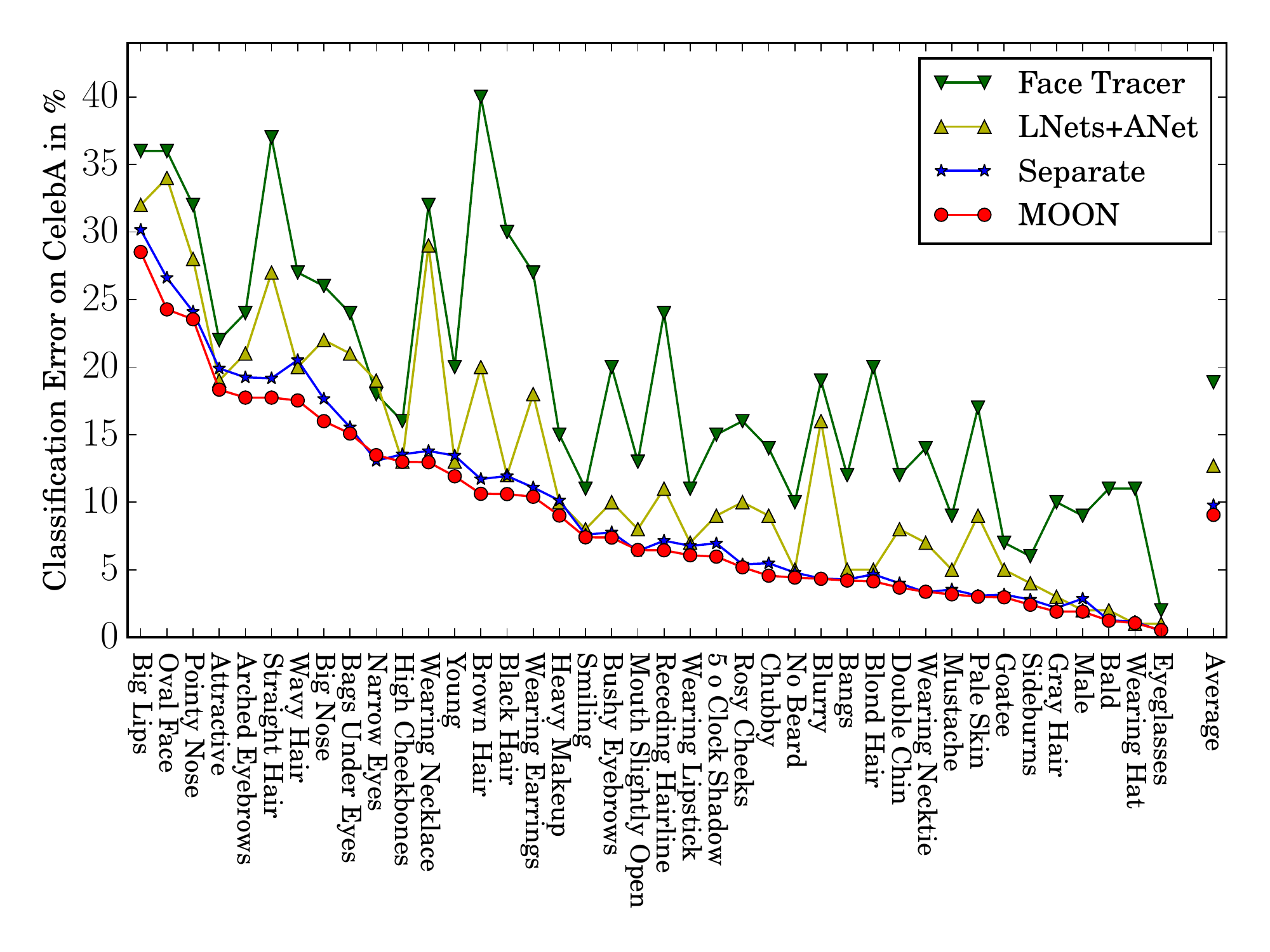}
  \cap{fig:Comparison}{Error Rates on CelebA}{This figure shows the classification errors on the test set of the CelebA dataset for several algorithms, including our Separate networks and MOON. The results of Face Tracer and LNets+ANet are taken from Liu et al.~\cite{liu2015deep}. For a tabular form of these results see the supplement to this paper.}
\end{figure*}

To compare with the results of Liu et al. \cite{liu2015deep}, we measure the success of our training in terms of classification error, 
 i.e., the number of cases, where our classifier $f$ predicted the incorrect label, relative to the total number of test images:
\begin{equation}
  \label{eq:classification_error}
  E_i({\bf X},{\bf Y}) = \frac1{N_{test}}\sum\limits_{j=1}^{N_{test}} e_i(X_j,Y_{ji}).
\end{equation}
The \texttt{Average} classification error is computed by taking the average of the classification errors over all ($M$) attributes:
\begin{equation}
  \label{eq:average_classification_error}
  \overline{E}({\bf X},{\bf Y}) = \frac1M\sum\limits_{i=1}^{M} E_i(X,Y).
\end{equation}
Note that this error does not differentiate between positive and negative values.
Hence, for very biased attributes, a random classifier which always predicts the dominant class would reach a low classification error, e.g., for \texttt{Bald} the random classification error would be as low as 2.24\,\%!

The classification errors for all the attributes are visually displayed in Fig.~\ref{fig:Comparison}.
There, we also included two results from Liu et al. \cite{liu2015deep}, converting from classification success (reported in \cite{liu2015deep}) to classification error.
The Face Tracer results reflect the best non-DCNN based algorithm that has been evaluated so far on the CelebA dataset.
LNets+ANet represent the state-of-the-art results on this dataset obtained by combining three different deep convolutional neural networks with support vector machines.

The average classification errors over all attributes for each classifier are: Face Tracer: 18.88\,\%, LNets+ANet: 12.70\,\%, Separate: 9.78\,\%, and MOON: 9.06\,\%. Thus, our MOON network achieves a relative reduction of 28.7\,\% of the error over the state of the art, and a 7.4\,\% reduction over the separately trained networks.
For almost all attributes, the results of our two approaches outperform the LNets+ANet state-of-the-art results, and the MOON network gives a lower error than the Separate networks trained specifically on a single attribute.

Interestingly, for several attributes that are traditionally not considered to be useful in face recognition, such as hair color (e.g. \texttt{Brown Hair}), hair style (e.g. \texttt{Straight Hair}), accessories (e.g. \texttt{Wearing Necklace}), and non face-related attributes (e.g. \texttt{Blurry}), our approach outperforms the LNets+ANet combination by an especially large margin.
We suspect that this effect is due to the fact that in \cite{liu2015deep}, the ANet network's feature space was derived from training on a \textit{face recognition} benchmark, and later adapted to the attribute classification task, which offers little direction for inferring the hidden representations of non facial identity related attributes.

\subsection{CelebAB: A Balancing Act}

\begin{figure*}[t!]
  \centering
  \subfloat[Unbalanced\label{fig:Distributions:Unbalanced}]{\includegraphics[width=.45\linewidth,page=1]{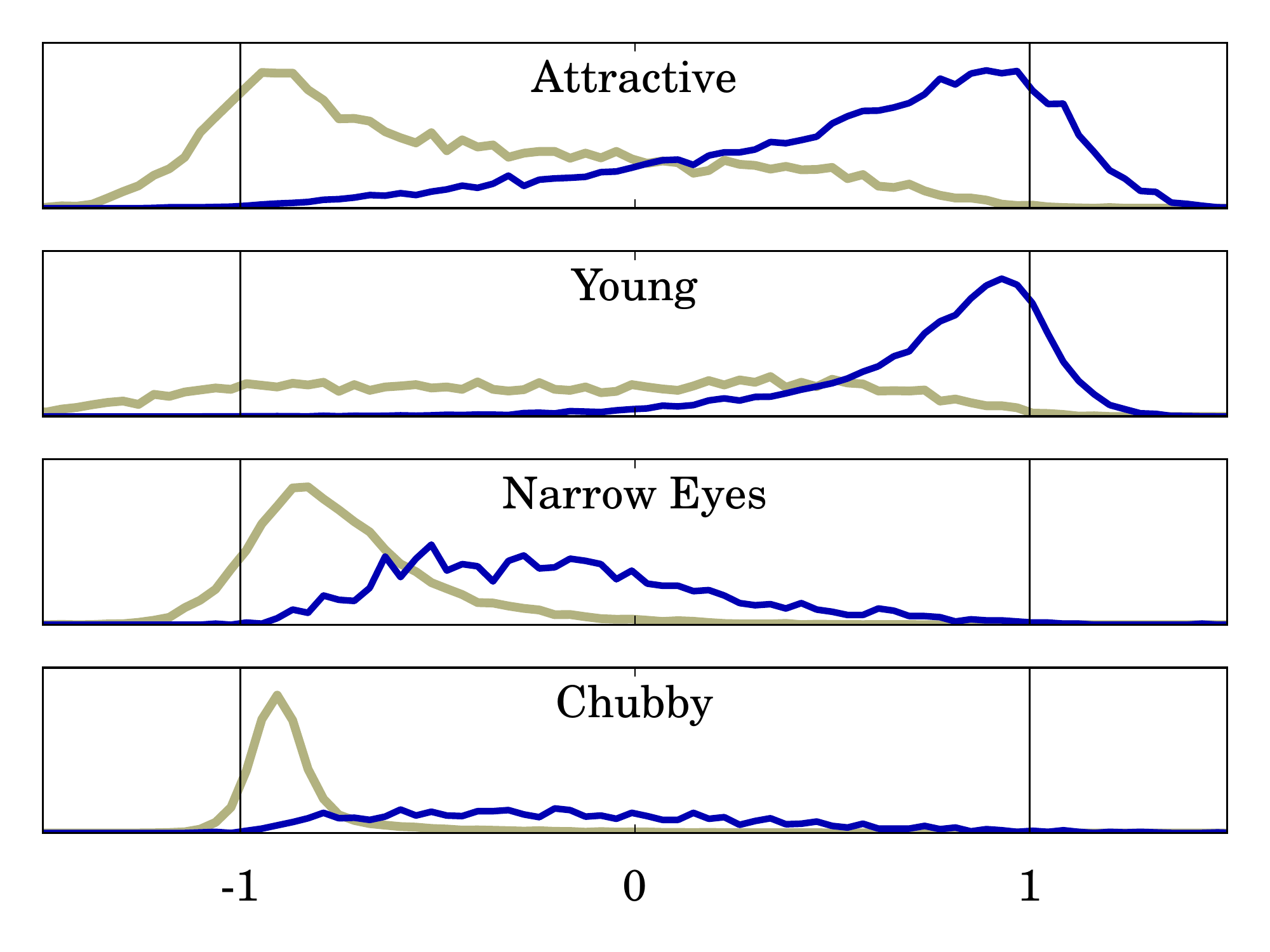}}%
  \hspace*{.03\linewidth}%
  \subfloat[Balanced\label{fig:Distributions:Balanced}]{\includegraphics[width=.45\linewidth,page=2]{Distributions.pdf}}
  \cap{fig:Distributions}{Score distributions}{This figure shows the distributions of the network outputs for four different attributes, when presenting images with present (blue) and absent (tan) attributes. In \protect\subref*{fig:Distributions:Unbalanced} network outputs after training with unbalanced data are shown, while in \protect\subref*{fig:Distributions:Balanced} the outputs of the network after training with the balancing loss layer are presented. Positive and negative score distributions are normalized independently.}
\end{figure*}

As demonstrated in Sec.~\ref{sec:celebA}, MOON obtains state-of-the-art classification accuracies on the CelebA dataset.
However, it is unclear how meaningful these results are for target distributions with different attribute frequencies, e.g., with more realistic distributions of \texttt{Young} or \texttt{Chubby} people.

Since our objective is to learn the network outputs to be $+1$ or $-1$ corresponding to presence or absence of attributes, respectively, we plotted the score distributions of the validation set for four of the attributes.
From Fig. \ref{fig:DatasetBias} we observe a strong bias for several attributes in the CelebA dataset, which we can find in the score distribution plots of Fig.~\subref{fig:Distributions:Unbalanced}, too.
Note that the positive and negative score distributions have been normalized independently, otherwise the positive scores for \texttt{Narrow Eyes} and \texttt{Chubby} would not be visible.
For attributes with a balanced number of positive and negative examples, such as \texttt{Attractive}, the distributions of negative (tan) and positive (blue) scores are also balanced.
On the other hand, for unbalanced attributes, such as \texttt{Young}, \texttt{Narrow Eyes} or \texttt{Chubby}, the dominant class is well distributed around its desired value, but the other class has not been learnt well. Interestingly, a comparably small bias in the training set (for \texttt{Young} there are 77\,\% positives and 23\,\% negatives) can destroy the capability of the network to learn the inferior class.

Intuitively, when having such unbalanced score distributions, one would expect that the threshold of 0 that we use for classification should be adapted.
However, given that the validation and test set follow the same bias as the training set, a threshold of 0 works well for the CelebA dataset.
Even more astonishingly, a wide range of thresholds around 0 will lead to approximately the same classification error and, hence, the network has learnt to balance between false positives and false negatives -- including the dataset bias.

To obtain balanced score distributions, we chose to have a balanced target distribution, i.e., $T_i^+=T_i^-=\frac12$ for each attribute $i$.
The resulting validation set score distribution for the same four attributes generated by the re-balanced MOON network after 34 training epochs can be seen in Fig.~\subref{fig:Distributions:Balanced}.
Apparently, the score distributions are much more balanced, and the threshold 0 seems to make more sense now.
Thus, one would expect that the classification error would be lower, too.
However, due to the high dataset bias, which is also present in the validation and test sets, the total average classification error of the balanced network on the (unbalanced) CelebA test set is 13.67\,\%.

Although this classification error is larger than that obtained by the unbalanced MOON network, \textit{this is an artifact of the significant imbalance in the original test set}; the error measure in Eq.~\eqref{eq:classification_error} has not been adapted to the target domain.
A fair comparison would measure the balanced classification error $E_i^B$ that weights the positive and negative classes according to the target distribution:
\begin{equation}
  \label{eq:adapted_error}
  E_i^B({\bf X},{\bf Y}) = \sum\limits_{j=1}^{N_{test}}
    \begin{cases}
      \frac{e_i(X_j,Y_{ji})T_i^+}{N_i^+} & \text{if } Y_{ji} = +1 \\[1.5ex]
      \frac{e_i(X_j,Y_{ji})T_i^-}{N_i^-} & \text{if } Y_{ji} = -1, \\
    \end{cases}
\end{equation}
where $N_i^+$ and $N_i^-$ are the respective numbers of positive and negative examples of attribute $i$ in the test set.
With $T_i^+=T_i^-=\frac12$, this error is effectively identical to the \emph{equal error rate} (EER) between errors made with positive and negative target values.
When computing classification error of the re-balanced MOON network example with $T_i^+=T_i^-=\frac12$, we obtain an average $E_i^B$ error of 12.98\,\%.

Note that the unbalanced MOON network, which is not trained to follow the target distribution, obtains an $E_i^B$ error of 21.41\,\%. This is precisely what we would expect of a domain adaptation system: A classifier adapted to the target distribution does better than a classifier that is not.

\section{Discussion}
\label{sec:discussion}

\subsection{Handling Mis-aligned Images}
\label{sec:misalignment}
In our experiments in Sec.~\ref{sec:experiments}, we used aligned images to train and test the networks.
To show that MOON is able to deal with badly aligned images, we conducted an additional experiment in which we used perturbed test images.
To perturb the images, we applied a random rotation within $\pm 10^{\circ}$, a random scaling with a scale factor in $\left[0.9,1.1\right]$, and a random translation of up to 10 pixels in either direction to the pre-aligned faces in the CelebA dataset.
We selected these parameters to be well outside of the error range of a reasonable (frontal face) eye detector.
Alignment errors of these magnitudes have been shown to \textit{highly} influence the performance of many traditional face recognition algorithms \cite{dutta14impact}.

When running this perturbed test set through our (unbalanced) MOON network, which was trained purely on aligned faces, we obtain a classification error of 11.62\,\%, which is higher than the 9.06\,\% obtained with aligned test images, but still better than the current state of the art in \cite{liu2015deep}.
We assume that we can improve the network stability against mis-alignment by incorporating augmented (e.g., misaligned perturbations) training data into the training process, since this has shown to improve the performance of DCNNs\cite{simard2003best}.

Some preliminary experiments seem to verify this claim: When training with mis-aligned and horizontally mirrored images (in total 10 copies for each training image), we were able to decrease the classification error on the mis-aligned test images to 9.50\,\%.
Unfortunately, this also caused a slight performance degradation when evaluating on purely aligned images, causing classification error to increase from the 9.06\,\% to 9.23\,\%.
Hence, in principle, the MOON architecture is able to work with aligned and mis-aligned images, as long as the conditions during training and testing are similar.
These tests further highlight the need to select data augmentation methods appropriate to the respective quality of the actual alignment algorithms used in real end-to-end systems.

\subsection{Face Verification on LFW}

One application of facial attributes is to enhance other recognition algorithms.
In order to evaluate our attribute classifiers on another dataset and to examine the effectiveness of our attributes for a particular application, we conducted the same View 2 LFW verification evaluation as Kumar et al. in ~\cite{kumar2009attribute}, using the 40 attributes extracted from MOON under both balanced and unbalanced networks.
We also tested the extracted attributes with respect to the features of Face Tracer (we downloaded the attribute vectors from \cite{kumar2009attribute} provided on the LFW web page \url{http://vis-www.cs.umass.edu/lfw}), using the approach detailed in \cite{kumar2009attribute}.
After optimizing RBF SVM parameters for each feature type separately using View 1 protocol of the labeled faces in the wild (LFW) dataset, the final classification accuracies that we obtained were $83.43\,\% \pm 2.22$ for Kumar's attributes, and $85.05\,\% \pm 1.57$ for the re-balanced MOON network.
Hence, our 40 MOON attributes provide better face recognition capabilities than the 73 attributes defined by Kumar et al.~\cite{kumar2009attribute}, though they are far from the current state-of-the-art on LFW.
This result, consistent with intuition, suggests that the accuracy of the attribute classification is important to providing noticeably better recognition results.
With $84.73\,\% \pm 1.99$, the verification accuracy for the unbalanced MOON network is only slightly lower than that of the re-balanced MOON network, but the stability is decreased.
We assume that training on a target distribution that better reflects the distribution of facial attributes in LFW will result in further increased accuracy/stability.
See the supplement to this paper for additional qualitative analysis of our LFW evaluation.

\section{Conclusion}

The MOON architecture achieves an accurate, computationally efficient, and compact representation which advances the state of the art on the CelebA dataset. 
Unlike competing approaches, our experiments did not rely on any datasets external to CelebA to train our network.
We also investigated dataset bias in CelebA and proposed domain adaptation methods for training to a different target distribution without requiring training samples from that population. 
Combining domain adaptive methods and multiple-task objectives into one mixed objective function, we conducted evaluations on a novel re-balanced version of CelebA (the \textit{CelebAB} dataset) and the LFW dataset that demonstrate the effectiveness of our approach.

Our work raises a philosophical question about the mathematics of attribute recognition: How \textit{should} the attribute recognition problem be treated?
Contrary to previous work, in which attribute labels are independently learnt, our approach implicitly leverages attribute correlations by explicitly forcing hidden layers in the network to incorporate information from multiple labels while simultaneously enforcing specified balance constraints via a domain adaptive loss. While CelebA labels are binary, MOON's weighted Euclidean loss also offers the capacity to learn labels along a continuous range, which is perhaps a more suitable representation for some attributes (e.g., \texttt{Big Nose},\texttt{Young}). Matching output score distributions to perceptual continuity and incorporating different types of attribute labels are interesting topics which we leave for future research.

\section*{Acknowledgments}
This research is based upon work  supported in part by the Office of the Director of National Intelligence (ODNI), Intelligence Advanced Research Projects Activity (IARPA), via IARPA R\&D Contract No. 2014-14071600012. The views and conclusions contained herein are those of the authors and should not be interpreted as necessarily representing the official policies or endorsements, either expressed or implied, of the ODNI, IARPA, or the U.S. Government. The U.S. Government is authorized to reproduce and distribute reprints for Governmental purposes notwithstanding any copyright annotation thereon.

\bibliographystyle{splncs}
\bibliography{paper}

\pagestyle{empty}


\section*{Supplementary material}

# *MOON*: A *Mixed Objective Optimization Network* for the Recognition of Facial Attributes Supplemental Material

Ethan M. Rudd, Manuel Günther, and Terrance E. Boult

Vision and Security Technology (VAST) Lab,
University of Colorado at Colorado Springs
{erudd,mgunther,tboult}@vast.uccs.edu

## 1   Hinge Loss Experiments

In our approach and experimental evaluation in the main text, we assigned attribute scores simply by thresholding the sign of the outputs from our network. In order to verify that there are no trivial accuracy gains at the output layer, we performed an additional SVM training on top of the 40-dimensional attribute vector output from the network. We took the finally selected (unbalanced) MOON network after 24 epochs, and extracted attribute vectors for the training, validation and test sets of CelebA. We trained 40 linear SVMs [1] on the training set, and used the validation set to optimize the C parameter for each attribute independently. Then we classified all extracted test set attributes. The final result was a classification error of 9.11 %, which is very close to, but still above, the 9.06 % that we obtained using the classification as given in Eq. (2) of the main text. Hence it seems that the MOON network has learnt a representation that is able to perform a multi-objective classification similar to the hinge-loss from Eq. (3) of the main text.

## 2   Qualitative Analysis on LFW

During our View 2 LFW evaluation on vectors of attributes, we recorded mis-classifications for each of the three approaches: Balanced MOON, Unbalanced MOON, and Face Tracer (the original approach from Kumar et al. [2]). After our View 2 evaluation, we performed an analysis of mis-classifications across all networks. Out of 6000 image pairs, balanced MOON mis-classified 1093, unbalanced MOON misclassified 1118, and Face Tracer mis-classified 1184. All three approaches mis-classified 470 common image pairs. We manually inspected these pairs to assess plausible reasons for the failures. We found that most of the mis-classified images for which identities matched (cf. Fig. 1(a)) had either 1.) tremendous photometric variations, 2.) objects occluding the face in one image and not in another, or 3.) different attributes for the same subject , e.g., in the top of Fig. 1(a), the capture subject has her hair dyed differently and is wearing much different shades of makeup and lipstick. For most of the mis-classifications

of non-match image pairs (e.g., Fig. 1(b)) the subjects tended to have similar facial attributes. This suggests that these mis-classifications on the View 2 protocol did not necessarily occur due to failures of the attribute classifiers themselves, but that the choice of attributes alone may not be sufficient to perform verification of certain image pairs.

Given two approaches, we can compare their similarity in terms of common error rate (CER). Given a set of errors $\mathbb{E}_1$ and a set of errors $\mathbb{E}_2$, the common error rate is given by:

$$CER(\mathbb{E}_1, \mathbb{E}_2) = \frac{\mathbb{E}_1 \cap \mathbb{E}_2}{\mathbb{E}_1 \cup \mathbb{E}_2}.$$

The unbalanced and balanced MOON networks mis-classified 755 images in common out of a total of 1456 mis-classified by either networks, yielding a CER of approximately 52 %. The balanced MOON and the Face Tracer approaches mis-classified 603 images in common out of 1674 mis-classified by either approach; a CER of 36 %. Unbalanced MOON and Face Tracer mis-classified 581 images out of 1721 in total, a lower CER of 34 %. The higher common mis-classification rate between the balanced MOON and Face Tracer are likely due to the implicit balance of the Face Tracer approach. Out of the 470 common mis-classifications the mis-classifications by all approaches, 294 were false positives while 176 were false negatives. This indicates a bias toward false positive classifications. Out of all pairs, 285 were mis-classified by both MOON approaches but correctly classified by Face Tracer, i.e., $|(\mathbb{E}_{Ba-MOON} \cap \mathbb{E}_{Un-MOON}) - \mathbb{E}_{FT}| = 285$. Out of these 285 errors, 182 were false positives while 103 were false negatives. Conversely, 470 images were mis-classified by Face Tracer but correctly classified by both MOON approaches, i.e., $|\mathbb{E}_{FT} - (\mathbb{E}_{Ba-MOON} \cup \mathbb{E}_{Un-MOON})| = 470$. Out of these 470 errors, 270 were false positives while 103 were false negatives.

Some examples of images in which MOON attributes from both balanced and unbalanced networks failed to offer successful verification but Face Tracer attributes succeeded are shown in Fig. 2. From our qualitative observations of this set of image pairs, we found that the subjects tended to be frontal facing especially for the match pairs – i.e., very few significantly off-pose images, and that both match and non-match pairs had disproportionate photometric, lighting, and skin color differences. We suspect that because the CelebA data, on which the MOON networks were trained had only 40 labeled attributes, with few corresponding to racial or photometric attributes (in contrast to Face Tracer), the choice of attributes from MOON networks as well as the learnt representation simply does not provide the information to disambiguate as effectively between skin color and photometric effects (e.g., glare).

Finally, some examples of images which were mis-classified when using Face Tracer attributes but correctly classified when using both MOON networks are shown in Fig. 3. In qualitative analysis we found that these images pairs often contained extreme pose differences between subjects. We hypothesize that MOON has learnt to deal with pose differences and mis-alignments much better than Face Tracer can – since Face Tracer relies on very precise alignment of

facial regions [2] – and for these images, Face Tracer simply fails to arrive at proper attribute scores.

# 3   Attribute Classification Error Rates for CelebA

For ease of comparison and reproducibility, we have tabularized the results displayed in Fig. 3 of the main text.

| Attribute | Face Tracer | LNets+ANet | Separate | MOON |
|---|---|---|---|---|
| 5 o Clock Shadow | 15.0 | 9.0 | 6.95 | **5.97** |
| Arched Eyebrows | 24.0 | 21.0 | 19.24 | **17.74** |
| Attractive | 22.0 | 19.0 | 19.9 | **18.33** |
| Bags Under Eyes | 24.0 | 21.0 | 15.55 | **15.08** |
| Bald | 11.0 | 2.0 | 1.26 | **1.23** |
| Bangs | 12.0 | 5.0 | 4.28 | **4.2** |
| Big Lips | 36.0 | 32.0 | 30.16 | **28.52** |
| Big Nose | 26.0 | 22.0 | 17.65 | **16.0** |
| Black Hair | 30.0 | 12.0 | 11.94 | **10.6** |
| Blond Hair | 20.0 | 5.0 | 4.66 | **4.14** |
| Blurry | 19.0 | 16.0 | 4.35 | **4.33** |
| Brown Hair | 40.0 | 20.0 | 11.71 | **10.62** |
| Bushy Eyebrows | 20.0 | 10.0 | 7.77 | **7.38** |
| Chubby | 14.0 | 9.0 | 5.48 | **4.56** |
| Double Chin | 12.0 | 8.0 | 4.0 | **3.68** |
| Eyeglasses | 2.0 | 1.0 | **0.47** | 0.53 |
| Goatee | 7.0 | 5.0 | 3.16 | **2.96** |
| Gray Hair | 10.0 | 3.0 | 2.18 | **1.9** |
| Heavy Makeup | 15.0 | 10.0 | 10.13 | **9.01** |
| High Cheekbones | 16.0 | 13.0 | 13.55 | **12.99** |
| Male | 9.0 | 2.0 | 2.86 | **1.9** |
| Mouth Slightly Open | 13.0 | 8.0 | **6.38** | 6.46 |
| Mustache | 9.0 | 5.0 | 3.54 | **3.18** |
| Narrow Eyes | 18.0 | 19.0 | **13.06** | 13.48 |
| No Beard | 10.0 | 5.0 | 4.79 | **4.42** |
| Oval Face | 36.0 | 34.0 | 26.61 | **24.27** |
| Pale Skin | 17.0 | 9.0 | 3.08 | **3.0** |
| Pointy Nose | 32.0 | 28.0 | 24.1 | **23.54** |
| Receding Hairline | 24.0 | 11.0 | 7.15 | **6.44** |
| Rosy Cheeks | 16.0 | 10.0 | 5.4 | **5.18** |
| Sideburns | 6.0 | 4.0 | 2.79 | **2.41** |
| Smiling | 11.0 | 8.0 | 7.59 | **7.4** |
| Straight Hair | 37.0 | 27.0 | 19.17 | **17.74** |
| Wavy Hair | 27.0 | 20.0 | 20.52 | **17.53** |
| Wearing Earrings | 27.0 | 18.0 | 11.09 | **10.4** |
| Wearing Hat | 11.0 | 1.0 | 1.15 | **1.05** |
| Wearing Lipstick | 11.0 | 7.0 | 6.77 | **6.07** |
| Wearing Necklace | 32.0 | 29.0 | 13.79 | **12.96** |
| Wearing Necktie | 14.0 | 7.0 | **3.33** | 3.37 |
| Young | 20.0 | 13.0 | 13.45 | **11.92** |
| Average | 18.88 | 12.7 | 9.78 | **9.06** |

Table 1: ERROR RATES ON CELEBA. *Percentage error rates on CelebA for for several algorithms, including our Separate networks and MOON. The results of Face Tracer and LNets+ANet (the previous state of the art) are adapted from Liu et al. [3], changing classification success to classification error. The best results are shown in bold.*

## Acknowledgments

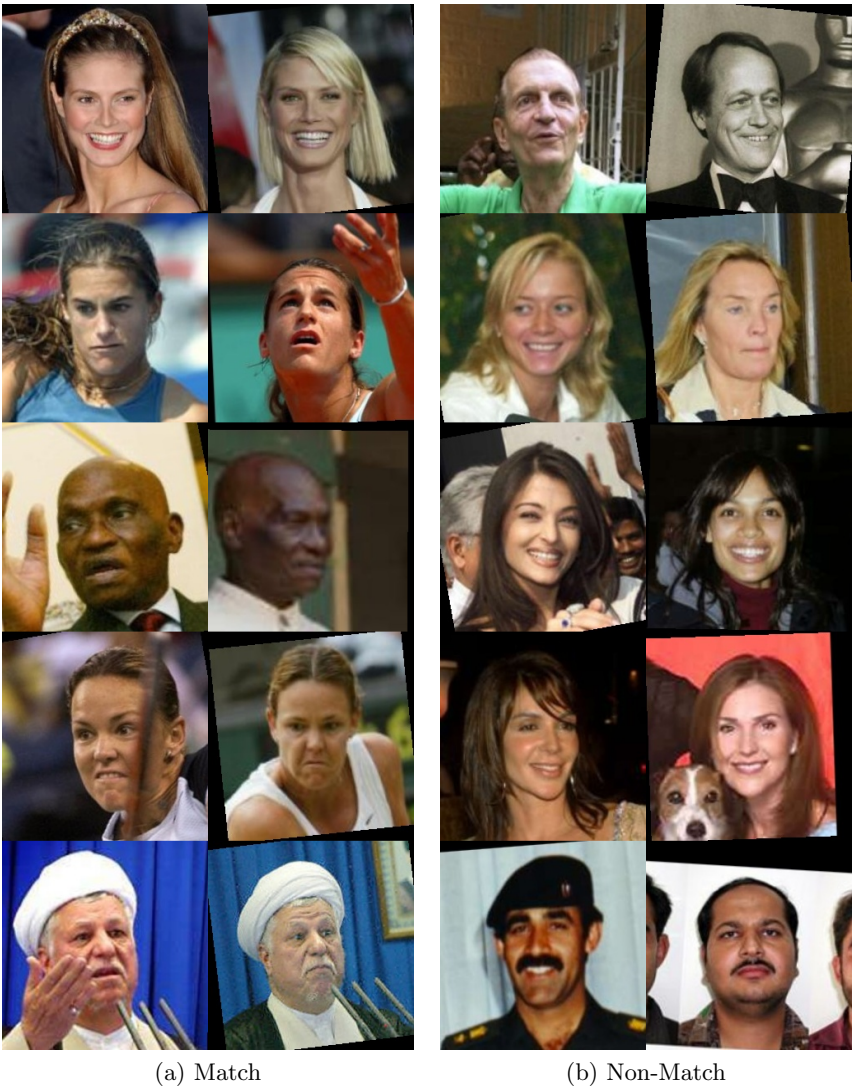

(a) Match                           (b) Non-Match

Fig. 1: MIS-CLASSIFIED BY ALL ALGORITHMS. *Examples of pairs of LFW images incorrectly classified when using attributes derived from any approach.*

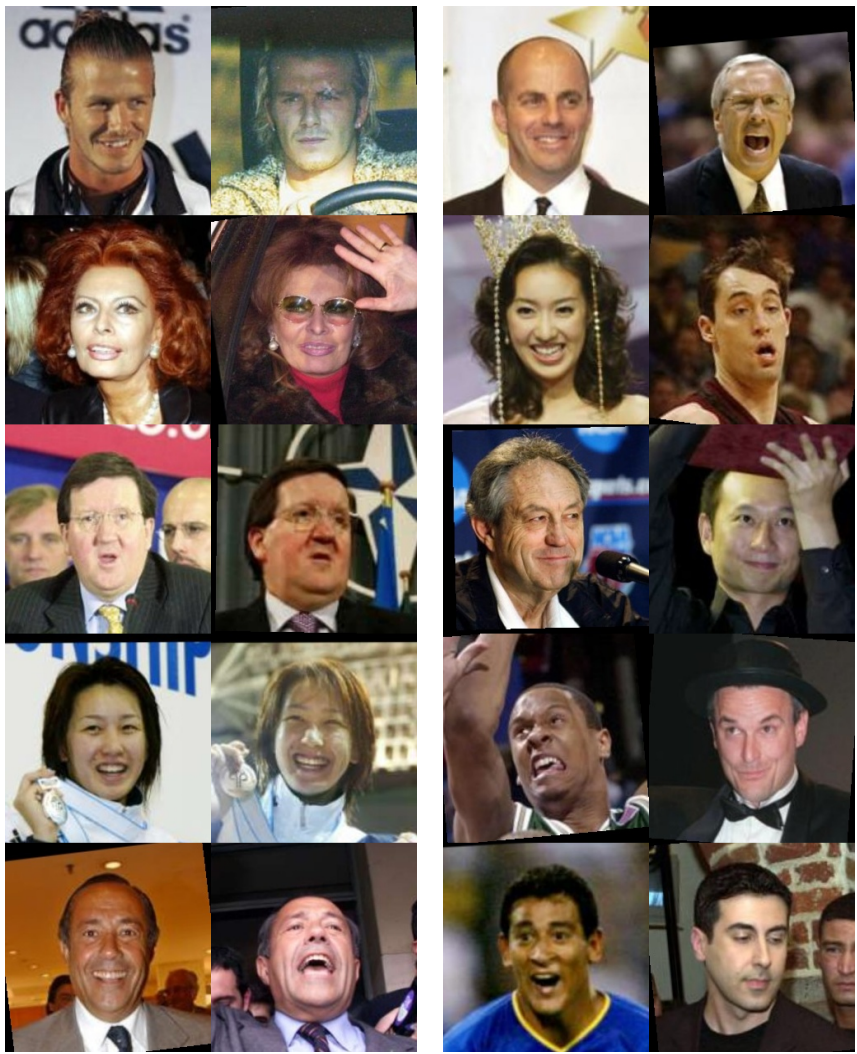

(a) Match                        (b) Non-Match

Fig. 2: MOON ERRORS. *Examples of pairs of LFW images that were correctly classified using Face Tracer attributes, but mis-classified using both balanced and unbalanced MOON attributes.*

(a) Match             (b) Non-Match

Fig. 3: FACE TRACER ERRORS. *Examples of pairs of LFW images that were incorrectly classified using Face Tracer attributes but correctly classified using attribute from both balanced and unbalanced MOON attributes.*


\end{document}